# Predicting cardiovascular risk from national administrative databases using a combined survival analysis and deep learning approach

Sebastiano Barbieri[1], PhD
Suneela Mehta[2],
Billy Wu[2],
Chrianna Bharat[3],
Katrina Poppe[2], PhD
Louisa Jorm[1], PhD
Rod Jackson[2], PhD

[1] Centre for Big Data Research in Health, University of New South Wales, Sydney, NSW, Australia
[2] Section of Epidemiology and Biostatistics, University of Auckland, Auckland, New Zealand
[3] National Drug and Alcohol Research Centre, University of New South Wales, Sydney, NSW, Australia

Corresponding author:
Sebastiano Barbieri, Centre for Big Data Research in Health, Level 2, AGSM Building (G27), UNSW Sydney, NSW 2052, Australia. E-mail: s.barbieri@unsw.edu.au

## Abstract

**Background**

Machine learning-based risk prediction models have the potential to outperform traditional statistical models in large datasets with many variables, by identifying both novel predictors and the complex interactions between them. This study compared the performance of deep learning extensions of survival analysis models with traditional Cox proportional hazards (CPH) models for deriving cardiovascular disease (CVD) risk prediction equations in national health administrative datasets.

**Methods**

Using individual person linkage of multiple administrative datasets, we constructed a cohort of all New Zealand residents aged 30-74 years who interacted with publicly funded health services during 2012, and identified hospitalisations and deaths from CVD over five years of follow-up. After excluding people with prior CVD or heart failure, sex-specific deep learning and CPH models were developed to estimate the risk of fatal or non-fatal CVD events within five years. The proportion of explained time-to-event occurrence, calibration, and discrimination were compared between models across the whole study population and in specific risk groups.

**Findings**

First CVD events occurred in 61,927 of 2,164,872 people. Among diagnoses and procedures, the largest 'local' hazard ratios were associated by the deep learning models with tobacco use in women (2·04, 95%CI: 1·99-2·10) and with chronic obstructive pulmonary disease with acute lower respiratory infection in men (1·56, 95%CI: 1·50-1·62). Other identified predictors (e.g. hypertension, chest pain, diabetes) aligned with current knowledge about CVD risk predictors. The deep learning models significantly outperformed the CPH models on the basis of proportion of explained time-to-event occurrence (Royston and Sauerbrei's R-squared: 0·468 vs. 0·425 in women and 0·383 vs. 0·348 in men), calibration, and discrimination (all $p<0·0001$).

**Interpretation**

Deep learning extensions of survival analysis models can be applied to large health administrative databases to derive interpretable CVD risk prediction equations that are more accurate than traditional CPH models.

**Funding**

New Zealand Health Research Council

# Introduction

Cardiovascular disease (CVD) risk equations, derived in clinical cohorts, are established means to inform clinical decisions regarding a person's CVD risk management.[1] They facilitate risk communication in a clinical setting and motivate adherence to recommended treatment and lifestyle modifications.[2] A complementary use of CVD risk equations is their derivation in routine administrative datasets and their application to every person in a given population for population health planning (e.g. estimation of future CVD incidence, identification of target sub-populations for prevention, and assessment of the likely benefit of health policies and interventions in different risk groups).[3,4] We have previously developed equations to estimate the five-year risk of a fatal or non-fatal CVD event across the primary prevention population of New Zealand, solely using linked routine national administrative health datasets, and these equations showed good calibration and discrimination across risk groups stratified by age, ethnicity, geographical region, level of deprivation and previous CVD-related pharmaceutical dispensing.[5]

CVD risk equations for population health planning differ from equations in clinical use as they can only consider predictors available in routinely collected administrative health data, which usually do not include smoking status, blood pressure and lipid profile. However, administrative health data may contain useful proxies for missing CVD predictors, e.g. diagnoses and procedures recorded during hospitalisations and pharmaceutical dispensing. If additional CVD predictors can be identified in routinely collected data, the predictive accuracy of CVD risk equations for population health planning can be improved. Whereas traditional methods for statistical inference on longitudinal data, such as Cox proportional hazards regression,[6] require predictors to be pre-specified and become less reliable as the number of predictors and possible associations among them increase,[7] machine learning can be used to identify relevant patterns across complex multimodal data. Recent methodological developments which replaced the linear combination of predictors in a Cox proportional hazards model with a deep neural network were able to demonstrate improved calibration and discrimination results.[8] In this study, deep learning extensions of survival analysis models were applied to routinely collected administrative health data to predict the five-year CVD risk of over two million adult New Zealanders.

This study had the following aims: (1) to develop novel deep learning models for predicting the five-year risk of a fatal or non-fatal CVD event across the New Zealand adult population without prior CVD or heart failure, using routinely collected administrative health data; (2) to determine which diagnoses, procedures, and dispensed medications are associated with increased risk of CVD event; (3) to compare the performance of the deep learning models and traditional Cox proportional hazards models on the basis of calibration and discrimination.

# Methods

**Study population and Data Sources**

The study population comprised New Zealand residents aged 30-74 years who were alive, in New Zealand, on December 31st, 2012 (index date) and with a health contact recorded in one or more of the following New Zealand routine national health databases: demographic characteristics, primary care enrolment (with voluntary re-enrolment occurring every three years), primary care visit reimbursements (to capture primary care visits by non-enrolled patients), community laboratory requests (but no laboratory results), community pharmaceutical dispensing, outpatient visits, hospitalisations and deaths. Data for each person were linked based on the National Health Index number (NHI number), a unique identifier assigned to every person who uses health and disability support services in New Zealand (estimated 98% of the population[9]). NHI numbers in the linked dataset were encrypted at source and all other individual patient identifiers were also removed. The dataset was linked to the Virtual Diabetes Registry, administered by the New Zealand Ministry of Health, to identify individuals with a history of diabetes as at December 31st, 2012. The age range for inclusion reflects the age group recommended for CVD risk assessment in New Zealand.[10]

People with a history of CVD or heart failure or missing predictor variables, were excluded from the dataset. People were considered to have a history of CVD or heart failure if their data contained relevant ICD10-AM codes associated with hospitalisations between January 1st, 1993 and the index date, if they were dispensed loop diuretics or antianginals three or more times in the five years prior to the index date, or if they were dispensed metolazone in the six months prior to the index date. The cohort development flowchart is presented in Figure 1. Additional information is reported in Supplemental Methods.

**Outcome**

The outcome of interest was the first fatal or non-fatal CVD event identified from national hospitalisation and mortality datasets over the five-year period between January 1st, 2013 and December 31st, 2017 (ICD10-AM codes are reported in Supplemental Methods). Five years is the recommended CVD risk prediction period in New Zealand guidelines.[10] People who died of CVD-unrelated causes during follow-up were censored. People who ceased to have any recorded health contact before December 31st, 2017 were assumed to have left New Zealand during follow-up and were also censored at their last recorded contact date with a health provider.

**Deep Learning Models**

*Predictors*

Deep learning models for predicting five-year CVD risk were developed using linked data for the described 2012 study population.[5] The input data contained both pre-specified predictors and all diagnoses, procedures, and medications in the five years prior to the index date.

Pre-specified predictors were included to facilitate comparisons between models and to ensure that the deep learning models had access to the information used by traditional Cox proportional hazards models. These pre-specified predictors were selected based on evidence regarding CVD risk factors and availability in the national health databases, and included: sex, age, ethnicity, level of deprivation, diabetes status, previous hospitalisation for atrial fibrillation (including both primary and secondary diagnoses between January 1st, 1993 and the index date) and baseline dispensing of blood-pressure-lowering, lipid-lowering and antiplatelet/anticoagulant medications, respectively. Deprivation was available in national health databases according to deciles of the New Zealand Index of Deprivation 2006 (NZDep2006), but was aggregated into quintiles (i.e. 1-5) to mitigate the effect of reassignment between deciles that occurs with different versions of the deprivation index over time. Age and deprivation quintiles were centred for analysis, using the mean value for age and the third quintile for level of deprivation. Changes in CVD pharmacotherapy over five years have been shown to be infrequent[11] and were not considered. First-order interaction terms were included based on clinical plausibility and statistical significance in traditional Cox proportional hazards models (*p*-value of <0·001).[11] Additional information regarding the pre-specified predictors is available in Supplemental Methods.

ICD10-AM coded diagnoses and procedures and dispensed medications were sorted chronologically by year and calendar month. Whenever a person's hospitalisation or medication period spanned multiple calendar months, the associated diagnoses, procedures, and medications were listed a corresponding number of times. Each listed ICD10-AM code was also associated with a variable indicating whether it was a primary diagnosis, a secondary diagnosis, an external cause of injury, or a procedure/operation. Rare ICD10-AM codes and medications associated with less than 500 people were excluded.

*Neural Network Architecture*

A schematic representation of the neural network used to map a person's pre-specified predictors, diagnoses, procedures, and medications to the log of the relative risk function is presented in Figure 2. At first, ICD10-AM codes and medications are represented as high-dimensional real-valued vectors (embeddings). The embeddings for ICD10-AM codes are summed with other embeddings describing the type of code (primary diagnosis, secondary diagnosis, external cause of injury, or procedure/operation). These vectors are then concatenated with a scalar value Δt, indicating the time difference, in months, between the current and the previous code in the clinical history. Next, the vectors are passed to three stacked bidirectional Gated Recurrent Unit (GRU) layers with 10% dropout. GRUs are a gating mechanism in recurrent neural networks: for each sequential input vector, a GRU outputs a vector which depends on the current input and the GRU's internal state (memory). Therefore, they are able to generate a vector representation of an input code in the context of a person's recent clinical history. The size of the GRU's hidden state was set equal to the input size. Three stacked layers were used to increase the network's expressive power. To focus on the most relevant outputs of the GRU layers and to obtain a single vector representation of the entire clinical history, a linear combination of the outputs was computed using dot-product attention. The resulting vector was concatenated with the pre-specified predictors, passed through a size-preserving fully-connected layer with exponential linear unit (ELU) activation and finally mapped to a scalar value (the log of the relative risk) by another fully-connected layer. Network hyperparameters were optimised using 10% of the data, stratified by outcome; details and additional references are reported in Supplemental Methods.

*Training*

The remaining 90% of the data were used to train and test the deep learning models, using stratified 5x2 cross-validation.[12] Sex-specific estimates of network parameters were obtained by maximising the Cox partial likelihood on the training data, using mini-batch stochastic gradient descent.[8] An Adam optimiser with a learning rate of 0·001 and β=(0·9, 0·999) was used for stochastic optimisation.[13] Training epochs iterated over all people who experienced a CVD event, and individual mini-batches consisted of 256 cases matched one-to-one with random controls in the respective risk set.[8] During hyperparameter optimisation, overfitting of the training data became apparent after approximately ten training epochs and therefore training was stopped after ten epochs. An ensemble of ten neural networks was constructed for each cross-validation fold by repeating training with different random parameter initialisations and averaging predictions.[14]

Five-year risk predictions were derived by estimating the baseline survival at the mean value of age, the third quintile of level of deprivation and the reference group of categorical variables.[15]

*Testing and Validation in New Zealand Sub-Populations*

The deep learning models were evaluated quantitatively based on the proportion of explained time-to-event occurrence (Royston and Sauerbrei's R-squared[16]), calibration (Brier score[17]), and discrimination (Harrell's C statistic[18] and Royston and Sauerbrei's D statistic[16]). Further qualitative assessment was performed through calibration plots and discrimination plots (i.e. dot charts of proportion of events occurring in each decile of predicted risk). Qualitative assessment was repeated for New Zealand sub-populations stratified by: (1) 15-year age bands; (2) ethnicity; (3) quintiles of deprivation; (4) dispensing of preventive medications.

*Statistical Inference*
To approximate the uncertainty of network parameters, the deep learning models were trained on the entire dataset 100 times with different random parameter initialisations.[14] For each trained model, a baseline log relative risk was computed for a person of mean age, in the third quintile of level of deprivation, in the reference group of categorical variables and with no associated diagnoses, procedures, or medications. The data for this person was then perturbed by either changing the values of continuous or binary predictors, or adding an individual diagnosis, procedure, or medication. The resulting change in log relative risk was used to estimate sex-specific 'local hazard ratios' (HRs) for the modified predictor. Local HRs averaged across trained models were reported together with 95% confidence intervals (CI). The local HRs are valid only for comparisons with the selected baseline population, whereas HRs derived by traditional Cox proportional hazards models describe general changes in hazard when other predictors are kept constant.

**Comparison with Cox Proportional Hazards Models**
Sex-specific, multivariable, Cox proportional hazards CVD risk models were developed using the same pre-specified predictors and first-order interaction terms used to develop the deep learning models. The methodology for developing the Cox models has been described previously in detail.[5] For the current study, these models were replicated in the 2012 New Zealand population. As for the deep learning models, calibration and discrimination measures were computed using 5x2 cross-validation. The statistical significance of differences between the deep learning and the traditional Cox models was assessed using combined 5x2 $F$ tests.[12] Hazard ratios were determined after fitting the Cox models to the entire dataset.

**Hardware and Software**
The deep learning models were implemented in Python 3.7.5 using PyTorch 1.5.1 and based in part on the PyCox library.[8] Hyperparameter optimization was performed using Optuna 1.5.0. One training epoch took approximately 90 seconds on a personal computer with a 3·50 GHz Intel Xeon processor, 64 GB of random access memory and a NVIDIA GeForce RTX 2080 Super graphics card. Statistical analyses to develop the Cox proportional hazards models were undertaken using Stata software version 14.1. The developed algorithms, trained deep learning models, and tabulated results are publicly available at https://github.com/VIEW2020/Varianz2012.

# Results
After applying the exclusion criteria (Figure 1), the cohort for this study comprised 2,164,872 New Zealand residents aged 30-74 years and still alive on December 31st, 2012 (Table 1). The proportion of women was 52·7%. The majority of the study population was of European (69·8% of women and 71·8% of men) and Māori (11·6% of women and 10·5% of men) descent, with 5·3% Pacific peoples, 3·5% Indian and 9·3% of other or unknown descent. The estimated prevalence of diabetes was around 6% for both sexes, and 0·6% of women and 1·2% of men had recorded diagnoses of atrial fibrillation. Blood-pressure-lowering medications were the most commonly dispensed category of baseline CVD preventive pharmacotherapy.

Among the women in this study, 2·1% experienced a CVD event during a mean follow-up time of 4·8 years (0·3% experienced a fatal CVD event). Among men, 3·7% experienced a CVD event during a mean follow-up time of 4·7 years (0·5% experienced a fatal CVD event).

**Deep Learning Models**
Tables 2 and 3 present the adjusted local hazard ratios for predictors in the sex-specific deep learning models, together with the proportions of people with each risk factor. Only the ten diagnoses and procedures and the ten medications associated with the largest local hazard ratios were reported (the full tables can be accessed at https://github.com/VIEW2020/Varianz2012).

The five-year risk of a first CVD event increased on average by 9% for each one year increment in age, for women and men. Event risk at any time during follow-up was greater among Māori and Pacific women and Māori, Pacific, and Indian men but lower among Indian and Other women and Other men compared with their European counterparts. Each increment in quintile of deprivation increased the CVD event risk by 16% in women and 10% in men.

Among the ICD10-AM codes, current tobacco use was associated with a doubling in CVD event risk in women (HR=2·04, 95% CI: 1·99 to 2·10) and an increase of 36% in men (CI: 31% to 41%). Codes related to essential hypertension, chest pain, diabetes, general anaesthesia for patients with severe systemic disease, chronic obstructive pulmonary disease, computerized tomography of the brain, history of long-term use of medications, retinopathy and retinal vascular changes, and chronic renal failure were associated with risk increases between 13% and 98% for both women and men. Hospital recorded alcohol use was associated with CVD event risk in women (45% risk increase, CI: 40% to 50%) but not in men (1% risk increase, CI: -1% to 4%). Some codes were associated with a 7 to 10% decrease in CVD event risk, such as childbirth in women and cycling injuries in men.

People with an increased risk of a CVD event were more likely to have been dispensed smoking cessation medications (nicotine, varenicline tartrate, buprorion hydrochloride), medications used for the treatment of raised blood pressure (cilazapril, furosemide, quinapril, felodipine, glyceryl trinitrate), bronchodilators (salbutamol with ipratropium bromide, tiotropium bromide) and statins (simvastatin). These findings were similar between women and men. Interestingly, dispensing of malathion (a head lice treatment) was also associated with increased CVD event risk in both women (37%, CI: 33% to 41%) and men (32%, CI: 28% to 36%).

**Comparison with Cox Proportional Hazards Models**
Both the deep learning models and the traditional Cox models showed good calibration and discrimination, for both women and men (Figure 3). However, the proportion of explained time-to-event occurrence was significantly larger for the deep learning models than for Cox models (0·468 vs. 0·425 in women and 0·383 vs. 0·348 in men; Table 4). Similarly, discrimination and calibration were significantly better for the deep learning models in terms of Royston and Sauerbrei's D statistic, Harrell's C statistic, and integrated Brier score, although differences were relatively small (Table 4). A qualitative evaluation of the calibration plots for sub-populations stratified by 15-year age bands, ethnicity, quintiles of deprivation and dispensing of preventive medications also suggested better calibration of the deep learning models (Supplemental Figures 1-14). Specific examples of differences in calibration between the two models in women and men aged 30-44 years, Māori women and men and most deprived women and men are shown in Figure 4. Overall, performance metrics associated with the models for women were better than those for men (Table 4).
Hazard ratios determined by the traditional Cox models were comparable in magnitude to the local HRs determined by the deep learning models (Table 5), although slightly smaller for ethnic groups and larger for history of diabetes and atrial fibrillation and baseline dispensing of medications. Coefficients of the corresponding CVD risk equations are reported in Supplemental Table 8.

# Discussion

This study developed deep learning models to predict the five-year risk of a fatal or non-fatal CVD event across the primary prevention population of New Zealand, using only predictors available in routinely collected administrative health data. The new models were used to gain insight into diagnoses, procedures and medications associated with increased risk of CVD events. Compared with traditional Cox proportional hazards models, the deep learning models showed improved calibration and discrimination across the whole population and in sub-populations stratified by 15-year age bands, ethnicity, quintiles of deprivation and dispensing of preventive medications.
A few previous studies investigated the use of machine learning for CVD risk prediction using large-scale data from prospective study cohorts,[19,20] family practices in the UK,[21] primary healthcare centres in Spain,[22] and hospitals and community health service centres in China[23] and the United States.[24] They suggest that machine learning improves CVD risk prediction, in agreement with the present findings, although their measures of performance were limited. Moreover, only one of these studies was able to account for censored data through the use of random survival forests.[19] A recent study comparing machine learning and traditional survival models for CVD risk prediction using UK family practice data shows that machine learning models which ignore censoring produce biased risk estimates, and suggests that survival models that consider censoring and that are explainable, are preferable.[25] However, the latter study did not evaluate any machine learning models that account for censoring. The present study is the first to use deep learning extensions of survival analysis models for CVD risk prediction, using routinely collected health data for a national population.
Previous studies generally used random forests to rank the importance of predictors, an approach which might not always be reliable due to bias towards inclusion of predictors with many split points.[26,27] In the present study, the associations of individual ICD10-AM codes and medications with the outcome were described using estimated hazard ratios. The most relevant diagnoses (e.g. current tobacco use, essential hypertension, chest pain, diabetes, chronic obstructive pulmonary disease, history of long-term use of medications, retinopathy and retinal vascular changes) and medications (related to smoking cessation, treatment of raised blood pressure and heart failure, bronchodilators, and statins) generally aligned with current knowledge about CVD risk predictors. The results also support previous findings regarding sex-related differences in cardiovascular risk predictors, such as the more deleterious effect of smoking in women and the particularly high risk associated with significant renal disease in men.[28]
The pre-specified predictors in the deep learning models were partly redundant: diabetes status, previous hospitalisation for atrial fibrillation and baseline dispensing of blood-pressure-lowering, lipid-lowering and antiplatelet/anticoagulant medications were included both as binary predictors and as individual ICD10-AM codes or medications. Similarly, first-order interaction terms between pre-specified predictors were part of the input but could also have been computed by the fully-connected layers of the deep learning models. This facilitated comparisons between models, but may also have biased the estimated local hazard ratios associated with the redundant predictors towards one. Since the deep learning models adjusted for any diagnoses, procedures, and medications in the five years prior to the index date, the estimated local hazard ratios might have been affected by collider bias and multicollinearity. Moreover, the hazard ratios reflect merely associations and not causal relationships between predictors and CVD risk, although frameworks that integrate deep learning and causal inference are being developed and represent an interesting venue for future research.[29]

In conclusion, the proposed deep learning extensions of survival analysis models enabled five-year CVD risk predictions for the primary prevention population of New Zealand with improved calibration and discrimination. The developed models are freely available and could similarly be used for CVD risk prediction and population health planning in other countries. The proposed method to compute local hazard ratios has applications beyond CVD risk prediction and could be used in time-to-event analyses to identify diagnoses, procedures, and medications associated with other conditions.

|  | **Women**[a] | **Men**[a] |
|---|---|---|
| **Participants** | 1,141,925 (52·7%) | 1,022,947 (47·3%) |
| **Age in years, mean (standard deviation)** | 49·0 (11·8) | 49·0 (11·6) |
| **Ethnicity** | | |
| European | 797,571 (69·8%) | 734,891 (71·8%) |
| Māori | 132,802 (11·6%) | 106,912 (10·5%) |
| Pacific | 60,965 (5·3%) | 54,659 (5·3%) |
| Indian | 38,481 (3·4%) | 36,248 (3·5%) |
| Other | 112,106 (9·8%) | 90,237 (8·8%) |
| **Deprivation quintile** | | |
| 1 | 272,564 (23·9%) | 242,794 (23·7%) |
| 2 | 244,140 (21·4%) | 216,602 (21·2%) |
| 3 | 227,684 (19·9%) | 202,118 (19·8%) |
| 4 | 212,257 (18·6%) | 190,774 (18·6%) |
| 5 | 185,280 (16·2%) | 170,659 (16·7%) |
| **Diabetes** | 67,143 (5·9%) | 65,290 (6·4%) |
| **Atrial fibrillation** | 6,393 (0·6%) | 11,900 (1·2%) |
| **Medications dispensed at baseline** | | |
| Blood-pressure-lowering | 194,670 (17·0%) | 167,839 (16·4%) |
| Lipid-lowering | 110,428 (9·7%) | 137,529 (13·4%) |
| Antiplatelet/anticoagulant | 64,158 (5·6%) | 79,443 (7·8%) |
| **Follow-up** | | |
| Total follow-up, years (mean) | 5,451,552 (4·8) | 4,792,390 (4·7) |
| CVD deaths | 2,986 (0·3%) | 5,153 (0·5%) |
| CVD events (non-fatal and fatal) | 23,592 (2·1%) | 38,335 (3·7%) |
| Median time to CVD event, years[b] (interquartile range) | 2·8 (1·4, 3·9) | 2·7 (1·4, 3·9) |
| Non-CVD deaths | 13,771 (1·2%) | 15,660 (1·5%) |

**Table 1.** Participant characteristics (N = 2,164,872). [a]Values are N (%) unless otherwise stated. [b]Among those with an event between 2013 and 2017 inclusively.

| **Women (N = 1,141,925)** | | **Deep learning model** |
|---|---|---|
| **Predictors** | **N (%)** | **Adjusted local HRs (95% CI)[a]** |
| **Age** (per year)[b] | | 1·09 (1·06, 1·11)[c] |
| | | |
| **Ethnicity** | | |
| New Zealand European | 797,571 (69·8%) | 1 |
| Māori | 132,802 (11·6%) | 1·96 (1·95, 1·97) |
| Pacific | 60,965 (5·3%) | 1·68 (1·67, 1·69) |
| Indian | 38,481 (3·4%) | 0·925 (0·918, 0·932) |
| Other | 112,106 (9·8%) | 0·720 (0·716, 0·723) |
| | | |
| **Deprivation quintile** (per quintile)[b] | | 1·16 (1·15, 1·16)[c] |
| | | |
| **Diabetes** | 67,143 (5·9%) | 1·39 (1·37, 1·40) |
| **Atrial fibrillation** | 6,393 (0·6%) | 1·68 (1·66, 1·69) |
| | | |
| **Medications dispensed at baseline** | | |
| Blood pressure lowering | 194,670 (17·0%) | 1·31 (1·29, 1·33) |
| Lipid lowering | 110,428 (9·7%) | 0·998 (0·990, 1·01) |
| Antiplatelet/anticoagulant | 64,158 (5·6%) | 1·46 (1·45, 1·47) |
| | | |
| **Interactions** | | |
| Age (years)*blood pressure lowering medication | | 0·980 (0·978, 0·982) |
| Age (years)*diabetes | | 0·999 (0·997, 1·00) |
| Age (years)*atrial fibrillation | | 0·963 (0·961, 0·966) |
| Blood pressure lowering medication*diabetes | | 1·10 (1·09, 1·11) |
| Antiplatelet/anticoagulant medications*diabetes | | 0·883 (0·874, 0·892) |
| Blood pressure lowering medication*lipid lowering medication | | 0·997 (0·989, 1·01) |
| | | |
| **Top-10 Diagnoses and Procedures** | | |
| Z72.0: Tobacco use, current | 84,589 (7·4%) | 2·04 (1·99, 2·10) |
| I10: Essential (primary) hypertension | 14,167 (1·2%) | 1·98 (1·91, 2·06) |
| R07.4: Chest pain, unspecified | 17,208 (1·5%) | 1·69 (1·63, 1·76) |
| 92514-39: General anaesthesia, ASA 3 (Patient with severe systemic disease that limits activity), nonemergency or not known | 10,961 (1·0%) | 1·55 (1·49, 1·61) |
| 56001-00: Computerised tomography of brain | 16,845 (1·5%) | 1·53 (1·47, 1·58) |
| J44.1: Chronic obstructive pulmonary disease with acute exacerbation, unspecified | 1,096 (0·1%) | 1·52 (1·47, 1·58) |
| Z92.2: Personal history of long-term (current) use of other medicaments | 2,661 (0·2%) | 1·52 (1·47, 1·58) |
| H35.0: Background retinopathy and retinal vascular changes | 692 (0·1%) | 1·51 (1·46, 1·57) |
| Z92.22: Personal history of long-term (current) use of other medicaments, insulin | 2,169 (0·2%) | 1·47 (1·42, 1·53) |
| Z72.1: Alcohol use | 957 (0·1%) | 1·45 (1·40, 1·50) |
| | | |
| **Top-10 Medications** | | |
| Nicotine | 79,506 (7·0%) | 1·74 (1·70, 1·78) |
| Varenicline tartrate | 31,750 (2·8%) | 1·54 (1·50, 1·58) |
| Furosemide [Frusemide] | 13,340 (1·2%) | 1·44 (1·40, 1·49) |
| Tiotropium bromide | 4,078 (0·4%) | 1·43 (1·39, 1·47) |
| Bupropion hydrochloride | 30,796 (2·7%) | 1·40 (1·36, 1·43) |
| Cilazapril | 76,762 (6·7%) | 1·38 (1·35, 1·41) |
| Malathion | 22,441 (2·0%) | 1·37 (1·33, 1·41) |
| Salbutamol with ipratropium bromide | 22,240 (1·9%) | 1·35 (1·32, 1·39) |
| Quinapril | 48,373 (4·2%) | 1·33 (1·30, 1·37) |
| Glyceryl trinitrate | 15,899 (1·4%) | 1·31 (1·26, 1·37) |

**Table 2.** Adjusted local hazard ratios (HRs) for time to CVD event within five years for women, determined by the deep learning model (only the 10 diagnoses and procedures, and the 10 medications, associated with the largest hazard ratios are reported). CI: confidence interval.

[a] The local hazard ratios for each predictor are adjusted for all other predictors. Values in parentheses are 95% confidence intervals unless otherwise stated.

[b] Age was centred at the mean value of 49·021. Deprivation quintile was centred around quintile three. The baseline survival estimate at five years for the deep learning model, relevant to the mean value of age, deprivation quintile three and the reference group of categorical variables was 0·9926104519395.

[c] Average and range (in parentheses) of estimated local hazard ratios for all values of the continuous predictor.

| Men (N = 1,022,947) | | Deep learning model |
|---|---|---|
| Predictors | N (%) | Adjusted local HRs (95% CI)[a] |
| **Age** (per year)[b] | | 1·09 (1·06, 1·13)[c] |
| | | |
| **Ethnicity** | | |
| New Zealand European | 734,891 (71·8%) | 1 |
| Māori | 106,912 (10·5%) | 1·69 (1·69, 1·70) |
| Pacific | 54,659 (5·3%) | 1·44 (1·43, 1·44) |
| Indian | 36,248 (3·5%) | 1·40 (1·39, 1·41) |
| Other | 90,237 (8·8%) | 0·785 (0·781, 0·790) |
| | | |
| **Deprivation quintile** (per quintile)[b] | | 1·10 (1·09, 1·10)[c] |
| | | |
| **Diabetes** | 65,290 (6·4%) | 1·46 (1·45, 1·47) |
| **Atrial fibrillation** | 11,900 (1·2%) | 1·61 (1·59, 1·62) |
| | | |
| **Medications dispensed at baseline** | | |
| Blood pressure lowering | 167,839 (16·4%) | 1·12 (1·11, 1·13) |
| Lipid lowering | 137,529 (13·4%) | 0·937 (0·929, 0·945) |
| Antiplatelet/anticoagulant | 79,443 (7·8%) | 1·43 (1·42, 1·44) |
| | | |
| **Interactions** | | |
| Age (years)*blood pressure lowering medication | | 0·987 (0·986, 0·989) |
| Age (years)*diabetes | | 0·993 (0·991, 0·995) |
| Age (years)*atrial fibrillation | | 0·994 (0·991, 0·996) |
| Blood pressure lowering medication*diabetes | | 0·969 (0·960, 0·978) |
| Antiplatelet/anticoagulant medications*diabetes | | 0·855 (0·848, 0·863) |
| Blood pressure lowering medication*lipid lowering medication | | 1·01 (1·01, 1·02) |
| | | |
| **Top-10 Diagnoses and Procedures** | | |
| J44.0: Chronic obstructive pulmonary disease with acute lower respiratory infection | 1,529 (0·1%) | 1·56 (1·50, 1·62) |
| N18.90: Unspecified chronic renal failure | 909 (0·1%) | 1·54 (1·49, 1·60) |
| R07.3: Other chest pain | 7,665 (0·7%) | 1·51 (1·45, 1·57) |
| E11.71: Non-insulin-dependent diabetes mellitus with multiple complications, stated as uncontrolled | 663 (0·1%) | 1·51 (1·45, 1·56) |
| L97: Ulcer of lower limb, not elsewhere classified | 896 (0·1%) | 1·50 (1·46, 1·55) |
| E11.72: Type 2 diabetes mellitus with features of insulin resistance | 6,209 (0·6%) | 1·50 (1·45, 1·55) |
| R07.4: Chest pain, unspecified | 15,470 (1·5%) | 1·47 (1·43, 1·52) |
| G62.9: Polyneuropathy, unspecified | 694 (0·1%) | 1·47 (1·41, 1·53) |
| Z92.2: Personal history of long-term (current) use of other medicaments | 2,336 (0·2%) | 1·47 (1·42, 1·53) |
| J44.9: Chronic obstructive pulmonary disease, unspecified | 523 (0·1%) | 1·46 (1·42, 1·51) |
| | | |
| **Top-10 Medications** | | |
| Quinapril | 46,541 (4·5%) | 1·73 (1·68, 1·78) |
| Varenicline tartrate | 26,037 (2·5%) | 1·73 (1·69, 1·76) |
| Nicotine | 64,493 (6·3%) | 1·68 (1·65, 1·71) |
| Simvastatin | 140,134 (13·7%) | 1·66 (1·62, 1·70) |
| Glyceryl trinitrate | 14,227 (1·4%) | 1·65 (1·58, 1·72) |
| Cilazapril | 79,241 (7·7%) | 1·60 (1·55, 1·64) |
| Bupropion hydrochloride | 25,139 (2·5%) | 1·58 (1·54, 1·61) |
| Tiotropium bromide | 3,399 (0·3%) | 1·52 (1·46, 1·58) |
| Salbutamol with ipratropium bromide | 14,745 (1·4%) | 1·46 (1·42, 1·49) |
| Felodipine | 38,670 (3·8%) | 1·39 (1·36, 1·43) |

**Table 3.** Adjusted local hazard ratios (HRs) for time to CVD event within five years for men, determined by the deep learning model (only the 10 diagnoses and procedures, and the 10 medications, associated with the largest hazard ratios are reported). CI: confidence interval.

[a]The local hazard ratios for each predictor are adjusted for all other predictors. Values in parentheses are 95% confidence intervals unless otherwise stated.
[b]Age was centred at the mean value of 49·027. Deprivation quintile was centred around quintile three. The baseline survival estimate at five years for the deep learning model, relevant to the mean value of age, deprivation quintile three and the reference group of categorical variables was 0·9812879278038.
[c]Average and range (in parentheses) of estimated local hazard ratios for all values of the continuous predictor.

| Performance metric | Statistic (95% CI) | | | | | |
|---|---|---|---|---|---|---|
| | Women | | | Men | | |
| | Deep learning | CPH | p-value[d] | Deep learning | CPH | p-value[d] |
| R-squared[a] | **0·468 (0·465, 0·471)** | 0·425 (0·423, 0·428) | <0·0001 | **0·383 (0·381, 0·385)** | 0·348 (0·346, 0·350) | <0·0001 |
| D statistic[b] | **1·92 (1·91, 1·93)** | 1·76 (1·75, 1·77) | <0·0001 | **1·61 (1·60, 1·62)** | 1·49 (1·49, 1·50) | <0·0001 |
| Harrell's C[b] | **0·813 (0·812, 0·814)** | 0·795 (0·794, 0·797) | <0·0001 | **0·771 (0·771, 0·772)** | 0·759 (0·758, 0·759) | <0·0001 |
| Integrated Brier score[c] | **0·00971 (0·00970, 0·00972)** | 0·00978 (0·00977, 0·00979) | <0·0001 | **0·0176 (0·0176, 0·0176)** | 0·0177 (0·0177, 0·0177) | <0·0001 |

**Table 4.** Performance metrics for the deep learning models and traditional Cox proportional hazards (CPH) models. Better results are in bold. 95% confidence intervals (CIs) are computed using 5x2 cross validation.

[a]Royston and Sauerbrei's R-squared measures how much of the time-to-event occurring is explained by the model. Higher values indicate that more variation is accounted for by the model.[16]

[b]Royston and Sauerbrei's D statistic and Harrell's C statistic are measures of discrimination. Better discrimination is indicated by higher values. Royston and Sauerbrei's D statistic represents the log hazard ratio of two equally sized prognostic groups identified by dividing the study population according to the median of the prognostic index. Therefore, the D statistic quantifies the prognostic separation of survival curves between these two groups.[16] Harrell's C statistic estimates the proportion of pairs of individuals where concordance is observed between predictions and outcomes.[18]

[c]The expected Brier score may be interpreted as the mean square error of prediction, and is affected by both calibration and discrimination.[17,30] Better discrimination and calibration is indicated by lower values. The integrated Brier score averages model performance over all available times.

[d]Computed using combined 5x2 $F$ tests.[12]

| Predictors | Adjusted hazard ratios (95% CI)[a] | |
|---|---|---|
| | **Women** | **Men** |
| **Age** (per year)[b] | 1·09 (1·09, 1·09) | 1·08 (1·08, 1·08) |
| | | |
| **Ethnicity** | | |
| European | 1 | 1 |
| Māori | 1·84 (1·78, 1·91) | 1·55 (1·51, 1·61) |
| Pacific | 1·40 (1·33, 1·48) | 1·26 (1·20, 1·32) |
| Indian | 0·910 (0·837, 0·989) | 1·18 (1·11, 1·25) |
| Other | 0·688 (0·647, 0·732) | 0·751 (0·717, 0·787) |
| | | |
| **Deprivation quintile** (per quintile)[b] | 1·15 (1·14, 1·16) | 1·11 (1·10, 1·12) |
| | | |
| **Diabetes** | 2·43 (2·26, 2·62) | 2·20 (2·07, 2·33) |
| **Atrial fibrillation** | 2·54 (2·14, 3·01) | 1·99 (1·80, 2·20) |
| | | |
| **Medications dispensed at baseline** | | |
| Blood pressure lowering | 2·24 (2·13, 2·35) | 1·86 (1·79, 1·94) |
| Lipid lowering | 1·02 (0·956, 1·08) | 0·942 (0·903, 0·982) |
| Antiplatelet/anticoagulant | 1·48 (1·42, 1·55) | 1·32 (1·27, 1·37) |
| | | |
| **Interactions** | | |
| Age (years)*blood pressure lowering medication | 0·975 (0·972, 0·978) | 0·976 (0·974, 0·979) |
| Age (years)*diabetes | 0·983 (0·980, 0·987) | 0·982 (0·979, 0·985) |
| Age (years)*atrial fibrillation | 0·984 (0·975, 0·994) | 0·985 (0·979, 0·991) |
| Blood pressure lowering medication*diabetes | 0·878 (0·807, 0·956) | 0·858 (0·803, 0·917) |
| Antiplatelet/anticoagulant medications*diabetes | 0·804 (0·744, 0·868) | 0·855 (0·803, 0·910) |
| Blood pressure lowering medication*lipid lowering medication | 0·858 (0·797, 0·923) | 0·941 (0·892, 0·994) |

**Table 5.** Adjusted hazard ratios for time to CVD event within five years, determined by the Cox proportional hazards models. CI: confidence interval.

[a]The hazard ratios for each predictor are adjusted for all other predictors.

[b]Age was centred in women and men separately using their mean values. For age, the mean value in women was 49·021 and the mean value in men was 49·027. Deprivation quintile was centred around quintile three in women and men. The baseline survival estimate at five years relevant to the mean value of age, deprivation quintile three and the reference group of categorical variables was 0·9905071151673 among women and 0·9782399916755 among men.

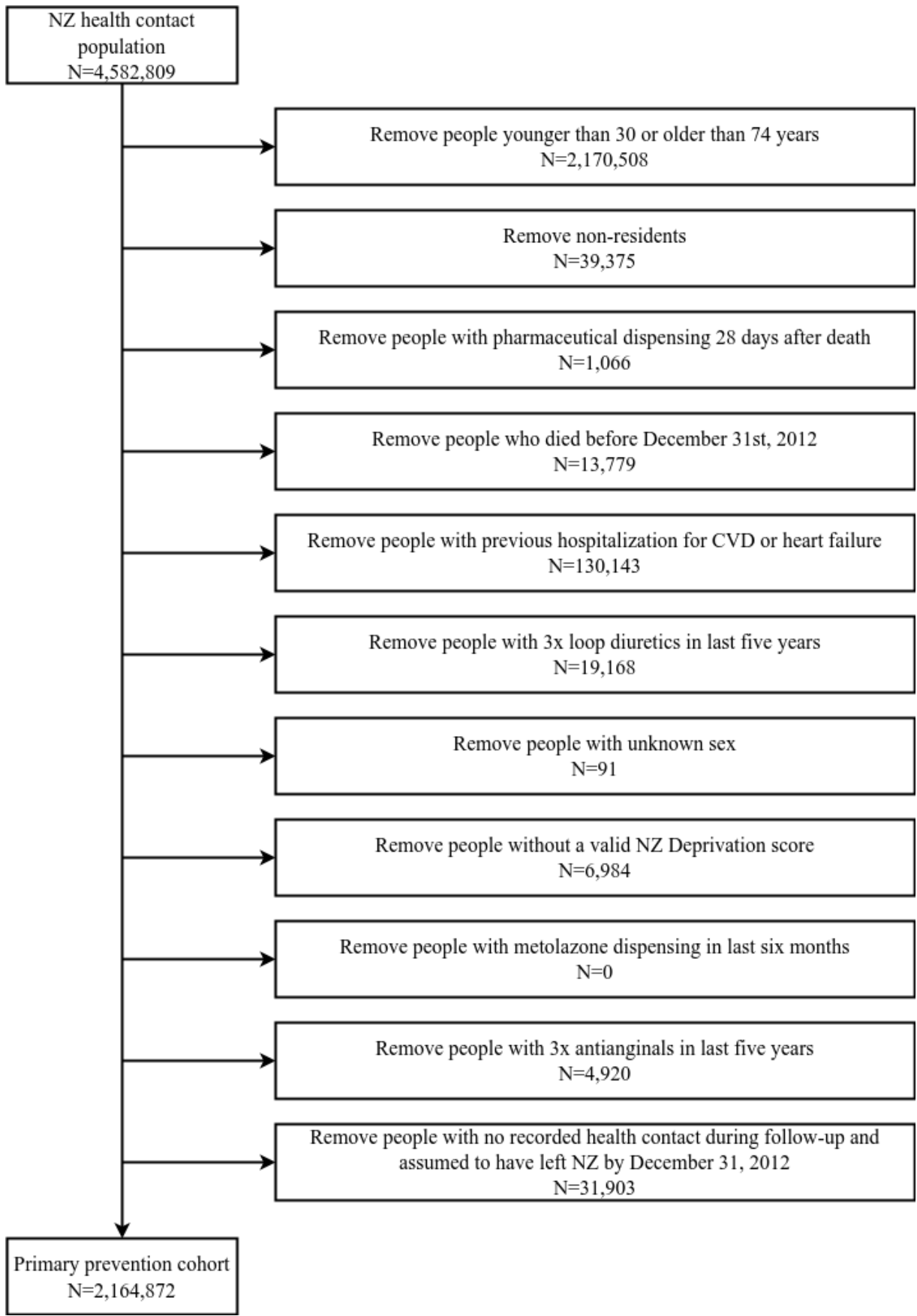

**Figure 1.** Cohort development flowchart.

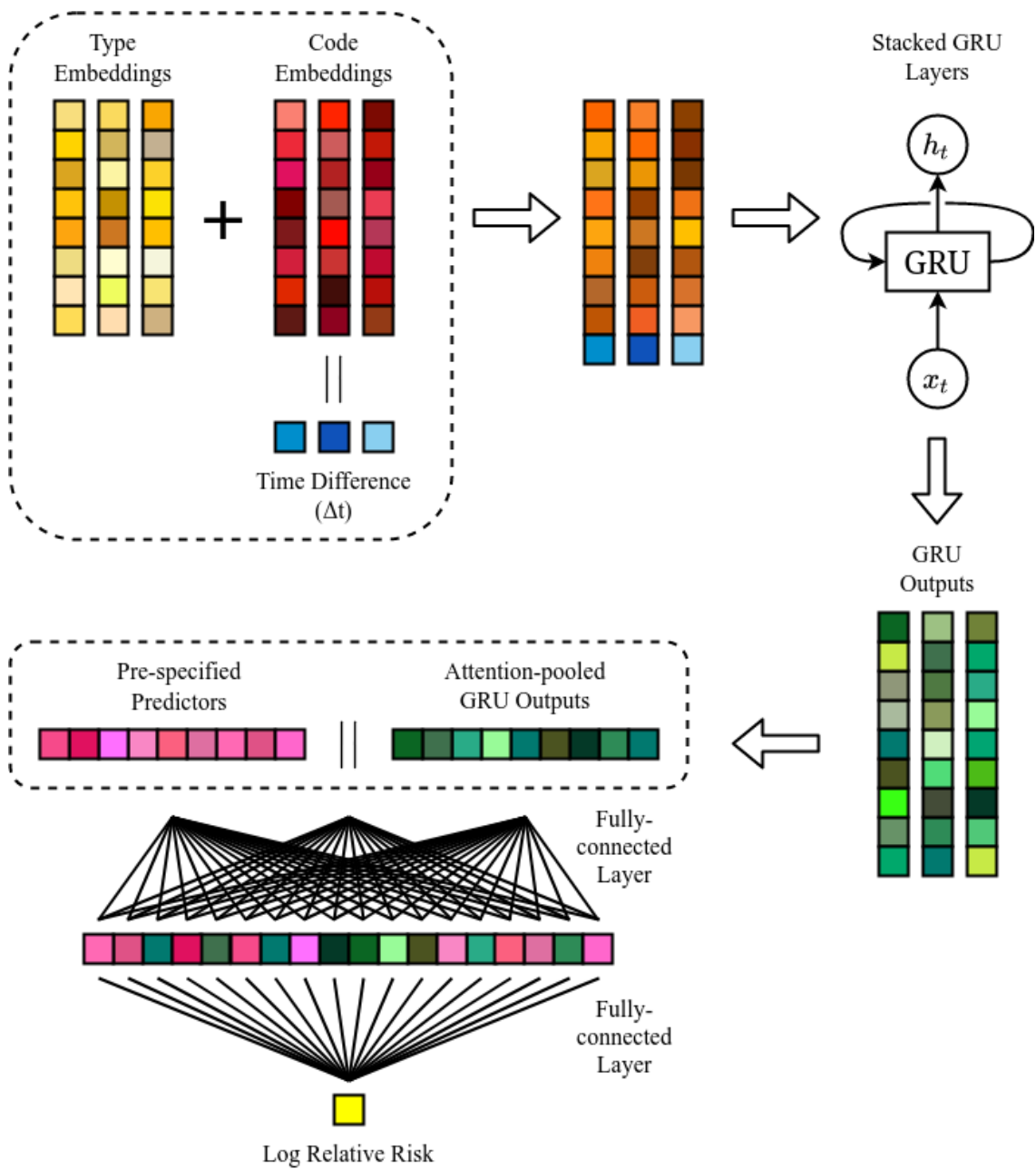

**Figure 2.** A schematic representation of the neural network used to map a person's predictors and clinical history to the log of the relative risk function. Code embeddings indicate vector representations of diagnoses, procedures and medications. Type embeddings describe the type of code (primary diagnosis, secondary diagnosis, external cause of injury, or procedure or operation). An extended description is reported in the main text. The '||' symbol indicates vector concatenation. Gated Recurrent Units (GRUs) are a gating mechanism in recurrent neural networks.

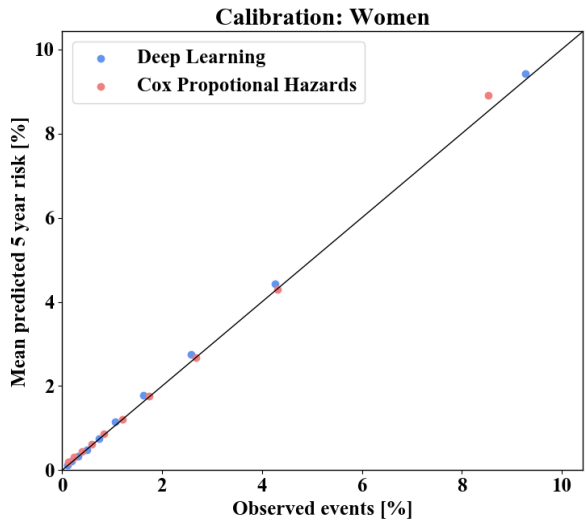
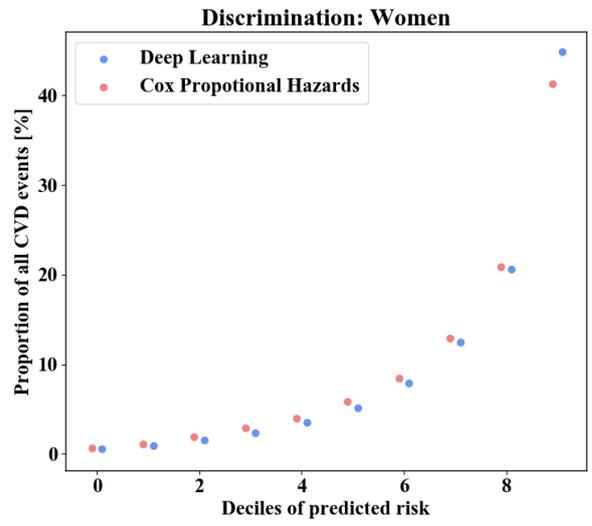
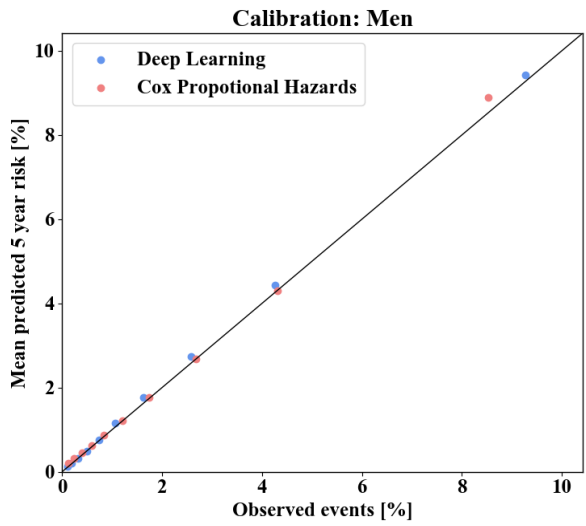
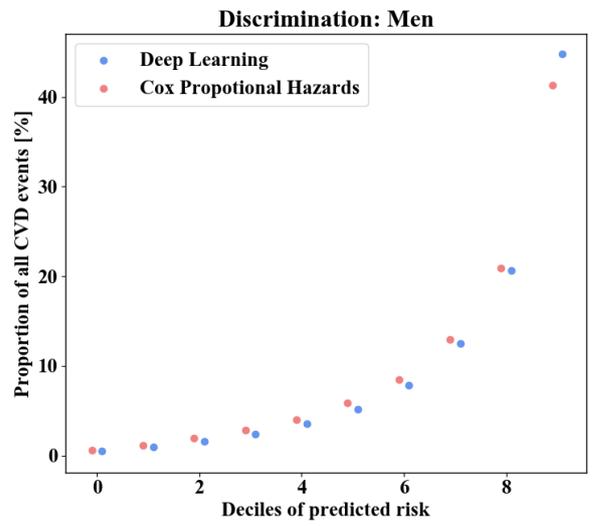

**Figure 3.** Calibration and discrimination of the deep learning models and Cox proportional hazards models for women and men. The calibration plots show the mean estimated five-year risk plotted against the proportion of CVD events that occurred over five years, for deciles of predicted risk. The diagonal line represents perfect calibration. The discrimination plots show the proportion of total observed events that occurred in each decile of predicted risk.

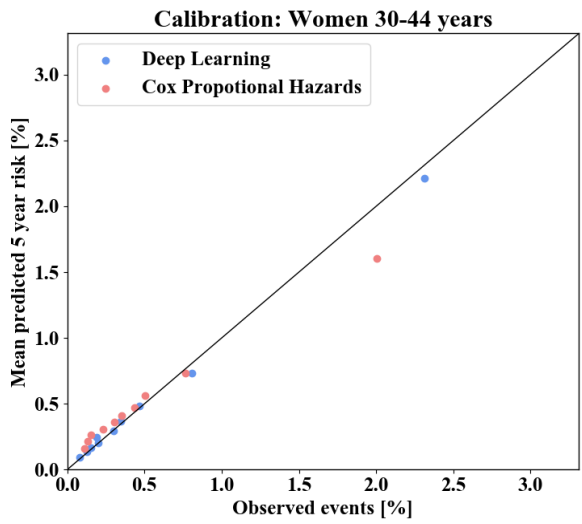
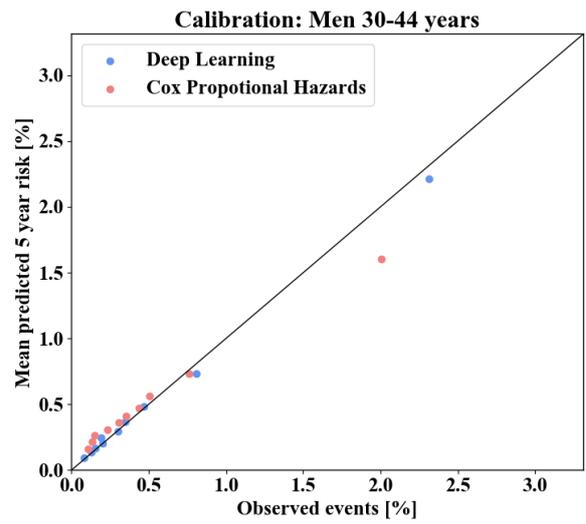
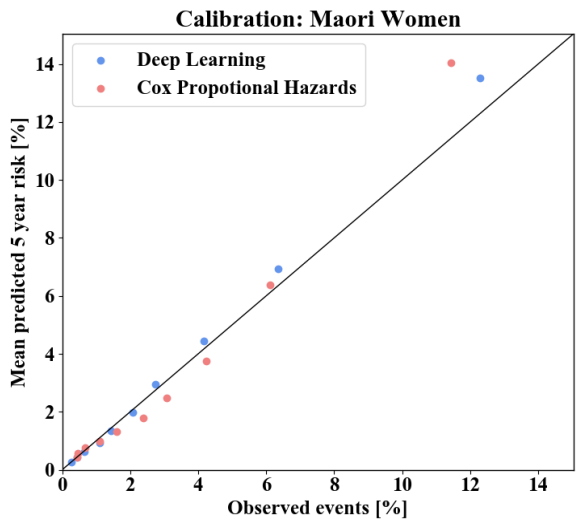
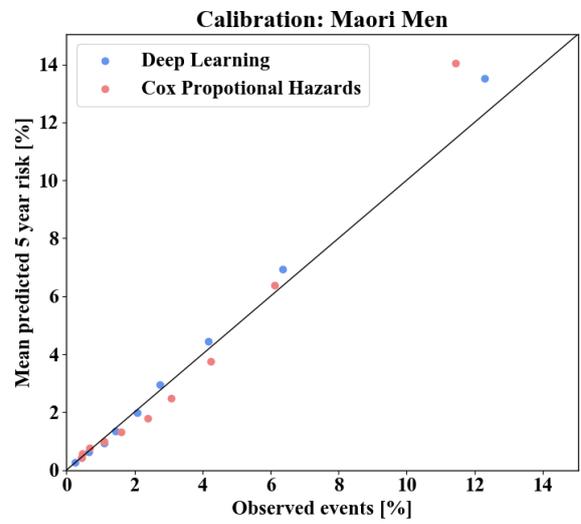
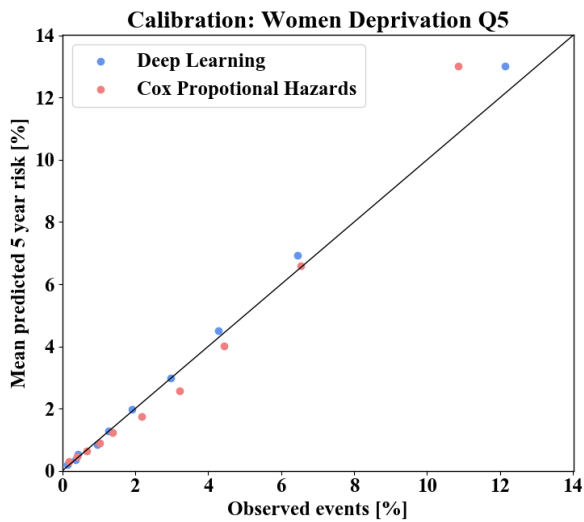
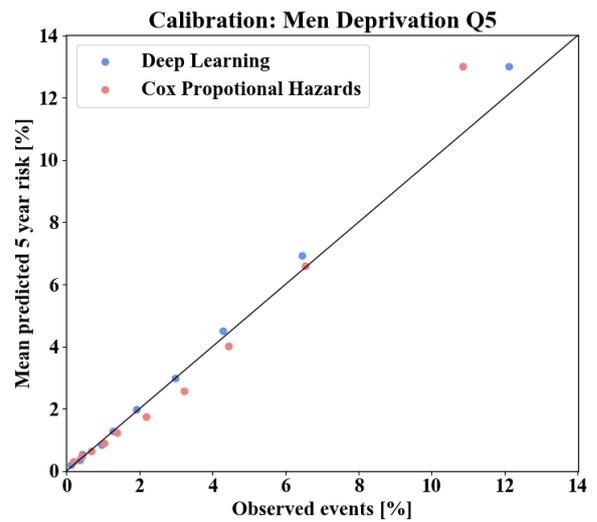

**Figure 4.** Calibration plots for the deep learning models and Cox proportional hazards models in specific New Zealand sub-populations (women and men aged 30-44 years, Māori women and men and most deprived women and men), suggesting improved calibration for the deep learning models.